\documentclass[10pt,twocolumn,letterpaper]{article}

 \usepackage[pagenumbers]{cvpr} 

\usepackage[dvipsnames]{xcolor}

\definecolor{cvprblue}{rgb}{0.21,0.49,0.74}
\usepackage[pagebackref,breaklinks,colorlinks,citecolor=cvprblue]{hyperref}

\usepackage{multirow}
\usepackage{amsmath}
\usepackage{dsfont}
\usepackage{pifont}

\newtheorem{lemma}{Lemma}

\def\paperID{} 
\def\confName{}
\def\confYear{}

\title{Unsupervised Multi-view Pedestrian Detection}

\author{Mengyin Liu\and Chao Zhu\thanks{Corresponding author.}\and Shiqi Ren\and Xu-Cheng Yin\and
	School of Computer and Communication Engineering, \\
	University of Science and Technology Beijing, Beijing, China \\
	{\tt\small blean@live.cn, shiqiren@xs.ustb.edu.cn, \{chaozhu, xuchengyin\}@ustb.edu.cn}\\
}

\begin{document}
\maketitle

\begin{abstract}
With the prosperity of the video surveillance, multiple cameras have been applied to accurately locate pedestrians in a specific area. However, previous methods rely on the human-labeled annotations in every video frame and camera view, leading to heavier burden than necessary camera calibration and synchronization. Therefore, we propose in this paper an Unsupervised Multi-view Pedestrian Detection approach (UMPD) to eliminate the need of annotations to learn a multi-view pedestrian detector via 2D-3D mapping. 1) Firstly, Semantic-aware Iterative Segmentation (SIS) is proposed to extract unsupervised representations of multi-view images, which are converted into 2D pedestrian masks as pseudo labels, via our proposed iterative PCA and zero-shot semantic classes from vision-language models. 2) Secondly, we propose Geometry-aware Volume-based Detector (GVD) to end-to-end encode multi-view 2D images into a 3D volume to predict voxel-wise density and color via 2D-to-3D geometric projection, trained by 3D-to-2D rendering losses with SIS pseudo labels. 3) Thirdly, for better detection results, i.e., the 3D density projected on Birds-Eye-View from GVD, we propose Vertical-aware BEV Regularization (VBR) to constraint them to be vertical like the natural pedestrian poses.  Extensive experiments on popular multi-view pedestrian detection benchmarks Wildtrack, Terrace, and MultiviewX, show that our proposed UMPD approach, as the first fully-unsupervised method to our best knowledge, performs competitively to the previous state-of-the-art supervised techniques. Code will be available.
\end{abstract}    
\section{Introduction}
\label{sec:intro}

\begin{figure}[t]
	\centering
	\includegraphics[width=1.0\columnwidth]{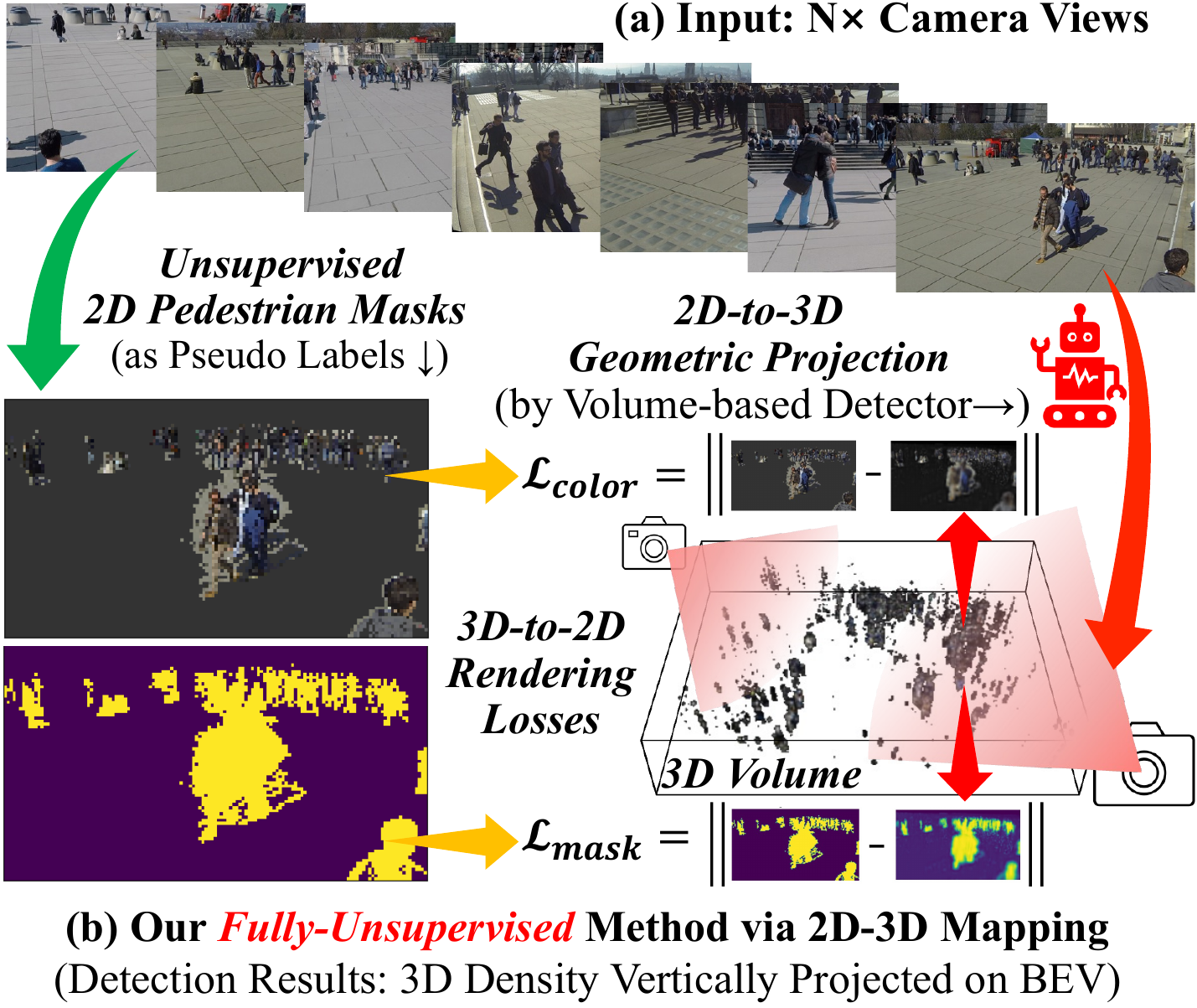}
	\caption{An overview of our proposed unsupervised pedestrian detection system via 2D-3D mapping. (a) The inputs are images from N$\times$ camera views. (b) Our fully-unsupervised method: unsupervised representations yield 2D pedestrian masks as pseudo labels. Volume-based detector encodes multi-view 2D images into a 3D volume, and learns to predict voxel-wise density and color from the  3D-to-2D rendering losses. 3D density is projected on Birds-Eye-View (BEV) as detection results, with high density values as pedestrian positions. }
	\label{Teaser}
\end{figure}

Detecting pedestrians is fundamental in various real-world applications, especially when the fine-grained positions of pedestrians on Birds-Eye-View (BEV) are required rather than coarse-grained bounding boxes \cite{liu2023vlpd}, such as crowd forecasting \cite{alahi2014socially} for safety and customer behavior analysis \cite{haritaoglu2013video} for retailing. To avoid occlusion or smaller scales, as shown in Figure \ref{Teaser}(a), multiple cameras around an interested region are introduced to capture pedestrian positions better. 

Most previous methods depend on supervised learning, including the classic detection-based \cite{xu2016multi, chavdarova2018wildtrack} and anchor-based detectors \cite{chavdarova2017deep, baque2017deep}, as well as more recent perspective-based anchor-free ones \cite{hou2020multiview, song2021stacked, hou2021multiview, qiu20223d, engilberge2023two}. They require supervised detectors or segmentation models, or map all 2D features onto BEV ground plane, but pedestrians land only on their feet. Thus, pedestrian BEV positions are labeled. 

\begin{figure*}
	\centering
	\includegraphics[width=0.975\textwidth]{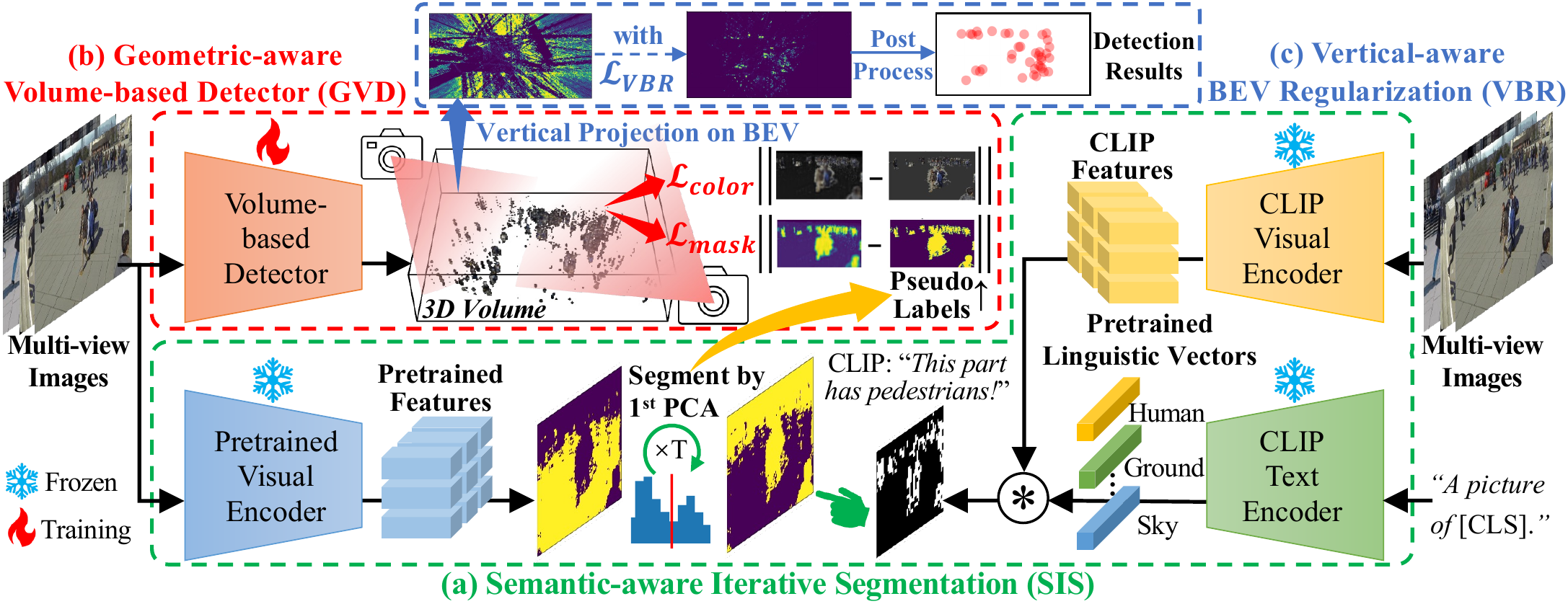}
	\caption{The architecture of our proposed UMPD approach. (a) Semantic-aware Iterative Segmentation (SIS) iteratively segments the PCA values of the unsupervised features into 2D masks as pseudo labels, equipped with the zero-shot semantic classes of CLIP \cite{radford2021learning} as a foreground selector. (b) Geometric-aware Volume-based Detector (GVD) encodes multi-view 2D images into a 3D volume via geometry, and learns to predict voxel-wise density and color by $\mathcal{L}_{\mathit{color}}$ and $\mathcal{L}_{\mathit{mask}}$ with SIS pseudo labels (cameras mean 3D-to-2D rendering). (c) Vertical-aware BEV Regularization (VBR) constraints the predicted 3D density to be ``human-like'' vertical on the BEV ground plane.}
	\label{Pipeline}
\end{figure*}

However, heavy labors are essential to create human-labeled datasets \cite{chavdarova2018wildtrack,fleuret2007multicamera} for the training of these detectors. On the one hand, typical raw data are video frames from multiple cameras, thus the data amount is more than an equal-length monocular video. On the other hand, annotators must carefully decide each pedestrian BEV position across multi-view images, plus to necessary camera calibration and synchronization. Recently, game engines are used to auto-generate images and labels \cite{hou2020multiview}, but suffer from various domain gaps, e.g., different color styles and camera poses. Therefore, annotating real data are still inevitable. 

In order to eliminate the need of annotations with heavy labors, as is shown in Figure \ref{Teaser}(b), we found a potential solution to locate pedestrians via 2D-3D mapping without any labels. Given the 3D existence (i.e., density) of pedestrians, the BEV labels mean a top-down observation, and their 2D masks from multiple camera views mean the surrounding observations. Hence, the 3D density becomes a ``bridge'', to be predicted by a volume-based detector from 2D multi-view images, learned from unsupervised 2D masks, and finally projected on BEV vertically as the detection results. 

For 2D pedestrian masks, we notice the recent powerful unsupervised pretrained models \cite{oquab2023dinov2,caron2021emerging,caron2020unsupervised,grill2020bootstrap}. Robust representations with inter-image co-exist concepts can be extracted by these models, and converted into the 1\textsuperscript{st} Principal Component Analysis (PCA) values to be segmented as foreground 2D masks. Meanwhile, unsupervised vision-language models like CLIP \cite{radford2021learning} can identify the zero-shot semantic classes with texts (e.g., ``\textit{A picture of a human}''). 

To construct a 3D volume from multi-view images, some methods for 3D object detection \cite{rukhovich2022imvoxelnet, tu2023imgeonet} or pose estimation \cite{iskakov2019learnable, nguyen2023deep} propose to firstly encode 2D features of each view, then back-project each pixel of feature into their potential 3D voxels based on their geometric correspondence via camera calibration, and finally fuse the volumes of each view into one. Although these methods are supervised, we are inspired by their powerful 3D volume construction frameworks to design a novel fully-unsupervised detector. 

Furthermore, differentiable rendering method \cite{johnson2020accelerating} can render the 3D density predicted by a volume-based detector into the 2D mask of each view, which can be learned from loss function with the 2D pseudo labels. To discriminate the crowded pedestrians from their appearances, the colors are also rendered and learned from the original 2D images. 

In conclusion, we have observed a high dependency of the current mainstream supervised methods on the laborious labels. As is illustrated in Figure \ref{Pipeline}, we propose a novel approach to tackle this problem via \textbf{U}nsupervised \textbf{M}ulti-view \textbf{P}edestrian \textbf{D}etection (\textbf{UMPD}). Our main contributions are: 

\begin{itemize}
	\item Firstly, Semantic-aware Iterative Segmentation (SIS) method is proposed to extract the PCA of the unsupervised representations, and segment them into 2D masks as pseudo labels. To identify the pedestrians, iterative PCA is used with zero-shot semantic classes of CLIP. 
	
	\item Secondly, we propose Geometric-aware Volume-based Detector (GVD) to encode multi-view 2D images into a 3D volume via geometry, and learn to predict 3D density and color from the rendering losses with SIS pseudo labels. Detection results are 3D density projected on BEV. 
	
	\item Thirdly, Vertical-aware BEV Regularization (VBR) method is further proposed to constraint the 3D density predicted by GVD to be vertical on the BEV ground plane, following the physical poses of most pedestrians.  
	
	\item Finally, formed by these key components, our proposed UMPD, \textbf{as the first fully-unsupervised method in this field to our best knowledge}, performs competitively on popular Wildtrack, Terrace, and MultiviewX datasets, especially compared with the previous supervised methods.
\end{itemize}
\section{Related Works}
\label{sec:related}

\subsection{Multi-view Pedestrian Detection}

To detect pedestrians in multi-view images, various methods have been proposed based on different architectures. 

Following the pedestrian detection for monocular image input, RCNN \& Clustering \cite{xu2016multi} firstly detects pedestrians by a RCNN detector \cite{girshick2014rich} in each single view, and then fuses these results across multiple views via clustering. Similarly, POM-CNN \cite{chavdarova2018wildtrack} relies on a segmentation model for single-view pedestrian masks that are fused as final prediction. In summary, fine-grained annotations are needed to learn the detector or segmentation models as their key components.

In an anchor-based style, DeepMCD \cite{chavdarova2017deep} predicts the pedestrian position on ground plane with their anchor box features. Deep-Occlusion \cite{baque2017deep} trains Conditional Random Field (CRF) as a high-order estimation on anchor features. To learn an anchor extractor, annotations are also needed. 

In an anchor-free style, MVDet \cite{hou2020multiview} firstly uses homography mapping from 2D visual features to a 2D BEV plane to predict pedestrian position as 2D Gaussians. Following MVDet, SHOT \cite{song2021stacked} projects on multiple heights. 3DROM \cite{qiu20223d} learns with a random occlusion augmentation. More augmentation is introduced by MVAug \cite{engilberge2023two}. MVDeTr \cite{hou2021multiview} predicts pedestrian shadow directions via Transformer \cite{vaswani2017attention}.

Differently, without any manual annotations, our proposed method UMPD performs 2D-3D mapping via predicting the 3D density of pedestrians via our volume-based detector GVD, learned from the 2D masks of our SIS as pseudo labels and our VBR as an extra regularization.

\subsection{Unsupervised Feature Representation}
\label{sec:UnsuperFeat}

In the past decade, unsupervised learning of feature representations has achieved powerful performance. Specifically, popular contrastive learning discriminates paired samples, which is capable of more zero-shot tasks than fine-tuning. 

For single modality, plenty of previous works \cite{caron2021emerging, oquab2023dinov2, caron2020unsupervised, grill2020bootstrap} learn representations as a no-label self-distillation, which uses more high-quality data and better contrastive learning techniques to learn robust visual representations, thus can filter 1\textsuperscript{st} PCA component as the foreground masks.

For multiple modalities, CLIP \cite{radford2021learning} learns the cross-modal mapping between vision and language, where positive samples are paired image and texts in the dataset, while negative ones are non-paired. Based on CLIP, MaskCLIP \cite{zhou2022extract} predicts 2D object masks queried by the input texts. 

In this paper, we propose to complement these powerful unsupervised models to obtain pseudo labels for Unsupervised Multi-view Pedestrian Detection (UMPD). In details, SIS is proposed to iteratively segment PCA of pretrained features for a fine-grained foreground segmentation, equipped with the zero-shot semantic capability from CLIP to identify the pedestrians as foreground from the segmented parts.

\subsection{Multi-view Construction of 3D Volume}

For various 3D tasks, it is crucial to construct an explicit 3D volume from multi-view images. For example, 3D pose estimation \cite{iskakov2019learnable, nguyen2023deep} assigns pixel-wise 2D features to their potential 3D positions as a volume for cross-view key-point correspondence. Some 3D object detectors \cite{rukhovich2022imvoxelnet, tu2023imgeonet} also construct an explicit volume to predict 3D bounding boxes. 

Another similar technique is Neural Radiance Field (NeRF) \cite{mildenhall2020nerf} which learns a neural network to implicitly represent a volume. With input ray and 3D voxel position, the outputs are density and color. However, typical $\le 10$ views \cite{chavdarova2018wildtrack,hou2020multiview} or even only $4$ views \cite{fleuret2007multicamera} in multi-view pedestrian detection datasets are difficult for NeRF to model the large-scale scenes with crowded pedestrians. The implicit volume is also inapt for vertical constraints like our VBR. 

To learn from 2D masks of our proposed SIS as pseudo labels, the differentiable rendering framework PyTorch3D \cite{johnson2020accelerating} renders an explicit 3D volume with the predicted density and color into 2D masks and images. Then, our proposed GVD detector predicts an explicit 3D volume and learns from 3D-to-2D rendering losses with pseudo labels. 

\section{Proposed Method}
\label{sec:method}

As is illustrated in Figure \ref{Pipeline}, our proposed UMPD approach comprises three key components: 1) Semantic-aware Iterative Segmentation (SIS) segments the PCA values into masks iteratively, with CLIP model to identify pedestrians; 2) Geometric-aware Volume-based Detector (GVD) encodes 2D images into a 3D volume with density and color, and learns from SIS masks by the rendering losses; 3) Vertical-aware BEV Regularization (VBR) method constraints the 3D density from GVD to be vertical on BEV. More details will be introduced in the following sections. 

\subsection{Semantic-aware Iterative Segmentation}
\label{sec:SIS}

With the powerful unsupervised methods \cite{oquab2023dinov2,caron2021emerging,caron2020unsupervised,grill2020bootstrap}, similar PCA values of pretrained features indicate the same concepts, which can be segmented into the 2D masks. Thus, we adopt them to identify pedestrians in multi-view images. 

\subsubsection{Unsupervised Segmentation of the 1\textsuperscript{st} PCA}

\begin{figure}[t]
	\centering
	\includegraphics[width=0.97\columnwidth]{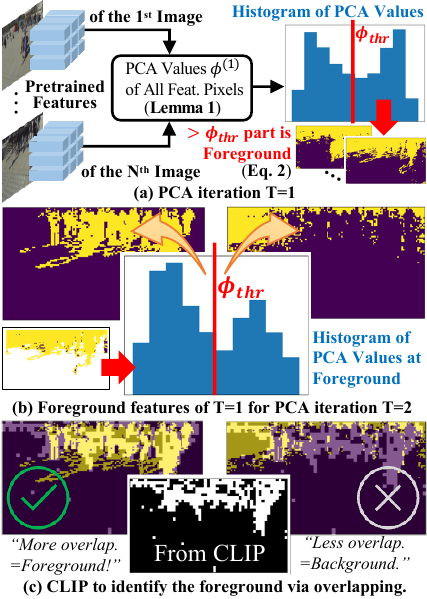}
	\caption{Our proposed Semantic-aware Iterative Segmentation (SIS). (a) Cross-image PCA values of pretrianed features are pixel-wisely segmented into a coarse-grained foreground. (b) Hence, foreground features are used for more iterations, with PCA values and their histograms in xy-axes. (c) Zero-shot semantic capability of CLIP \cite{radford2021learning} identifies the foreground via overlapping. }
	\label{SIS}
\end{figure}

Given $\mathrm{N}\times$ multi-view images, their pretrained features are  $\mathbf{X}=(X_1^\top, X_2^\top,\cdots,X^\top_\mathrm{P})^\top $, where $\mathrm{P=N\times H \times W}$ is total pixels and each $X_p \in \mathbb{R}^\mathrm{D}$. Denote $Z$ as a linear mapping of $\mathbf{X}$ by the vector $\phi = (\phi_{1}, \phi_{2},\cdots,\phi_\mathrm{P})^\top \in \mathbb{R}^\mathrm{P}$, where $\sum_{d=1}^\mathrm{D} x_{pd} = 0$ after the Zero Standardization of each $X_p$:
\begin{equation}
	Z = \phi_{1}X_1 + \phi_{2}X_2 + \cdots + \phi_\mathrm{P}X_\mathrm{P}.
	\label{PCA-1}
\end{equation}

\begin{lemma}
	\label{lemma:1}
	The 1\textsuperscript{st} PCA vector $\phi^{(1)}$ of $\mathbf{X}$: {\rm(a)} maximizes the global variance $\mathrm{var}(Z^{(1)}) = \frac{1}{\mathrm{D}}\sum_{d=1}^\mathrm{D} (z^{(1)}_{d})^2$; {\rm(b)} minimizes the reconstruction loss $\min_{\mathbf{W}} \Vert \mathbf{X} - \mathbf{W} (\mathbf{W}^\top \mathbf{X}) \Vert^2$, where $\mathbf{W} = (W_1^\top, W_2^\top,\cdots,W_\mathrm{P}^\top)^\top \in \mathbb{R}^\mathrm{P\times D}$ replaces $\phi^{(1)}$ for a bi-direction mapping between $Z^{(1)}$ and $\mathbf{X}$.
\end{lemma}

According to the bi-direction mapping in \cref{lemma:1}, for the foreground concepts that co-exist in multiple images, e.g., pedestrians, if their features $\mathbf{X}$ are similar, their 1\textsuperscript{st} PCA values $\phi^{(1)}$ are also similar, as is shown in the histogram of Figure \ref{SIS}(a). Then,  $\phi^{(1)}_{p} \in \mathbb{R}$ can be more easily divided by a threshold than the complex vectors $X_p \in \mathbb{R}^\mathrm{D}$. Formally, given the pixel-wise PCA values $\Phi_{nij}$ of $\mathbf{\Phi} \in \mathbb{R}^\mathrm{N\times H \times W}$,  where $\mathbf{\Phi}$ are reshaped from $\phi^{(1)} \in \mathbb{R}^\mathrm{P}$, they are segmented into masks $\mathbf{M}^{t=1}_n$ of each view $n$ in Figure \ref{SIS}(a) via the threshold value $\phi_\mathit{thr}$, which is formulated as: 
\begin{equation}
	\mathbf{M}^{t=1}_{nij} =\left\{
	\begin{aligned}
		1.0 & , & \Phi_{nij} > \phi_\mathit{thr}, &\mathrm{~~as~foreground}, \\
		0.0 & , & \Phi_{nij} \leq \phi_\mathit{thr}, &\mathrm{~~as~background}. \\ 
	\end{aligned}
	\right.
	\label{PCA-naive}
\end{equation}

\begin{figure*}
	\centering
	\includegraphics[width=0.985\textwidth]{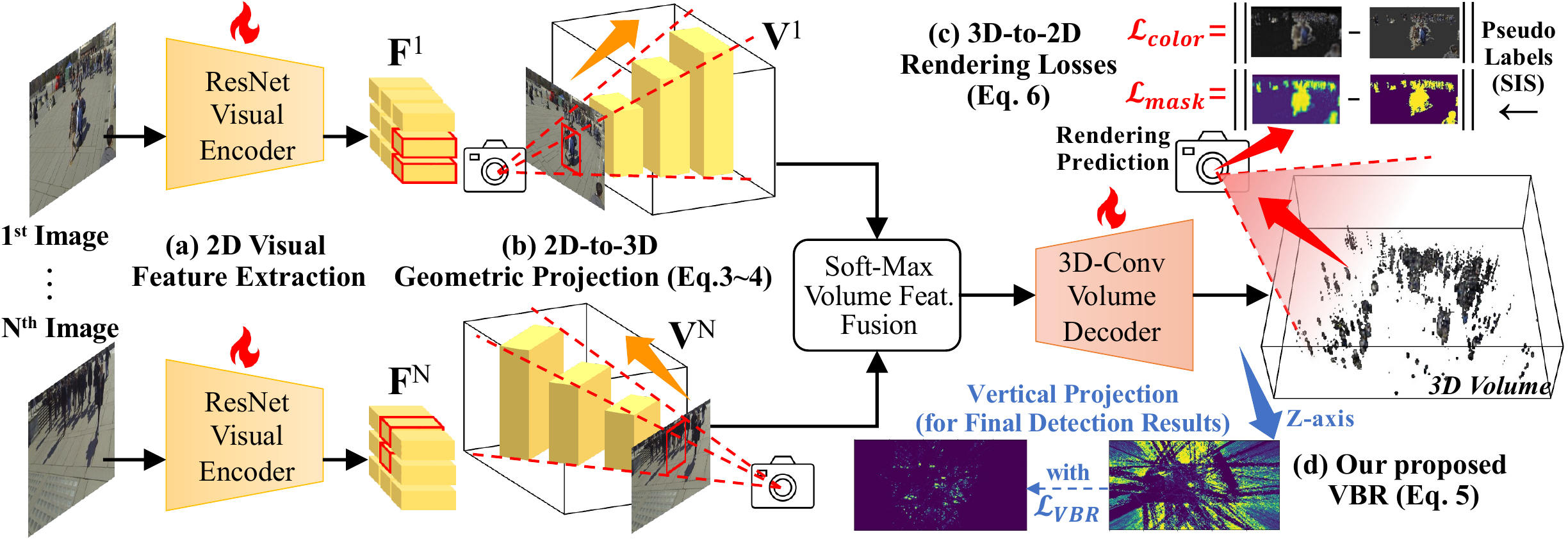}
	\caption{Our proposed Geometric-aware Volume-based Detector (GVD) and Vertical-aware BEV Regularization (VBR). (a) 2D features $F^n$ of each image view $n$ are extracted by a visual encoder. (b) 2D-to-3D geometric projection assigns the pixel-wise features (take the highlighted red-border features as examples) to its potential 3D voxel positions via Eq. \ref{Backproj} and \ref{Assign} as 3D volumes. (c) 3D-to-2D rendering losses are equipped with PyTorch3D framework \cite{johnson2020accelerating}, which renders the predicted 3D density and color into 2D masks and images, and 2D masks from SIS serve as pseudo labels. (d) Our proposed VBR constraints on vertical direction for better detection results.}
	\label{GVD}
\end{figure*}

\subsubsection{Semantic-aware Iterative Segmentation of PCA}
\label{sec:SIS-PCA}
As is illustrated in the Figure \ref{SIS}(a), due to the complex realistic scenarios, the visualized PCA segmentation is a mixture of pedestrian and other objects. Recalling \cref{lemma:1}, unsupervised features are not fully ideal. Therefore, we propose multiple iterations of PCA to tackle this problem. In details, Given $\mathbf{M}^{t}$, we calculate PCA again from its forground features $\mathbf{X}^{t+1} = \mathbf{X}^{t}[\mathds{1}(\mathbf{M}^{t} = 1.0)] \in \mathbb{R}^\mathrm{P'\times D}, \mathrm{P' \ll P}$. However, in Figure \ref{SIS}(b), the $\leq \phi_\mathit{thr}$ part is the real foreground, which is inconsistent with Eq. \ref{PCA-naive}.

Inspired by the zero-shot capability to recognize the object classes from vision-language pretrained model like CLIP \cite{radford2021learning}, cosine similarities between pretrained linguistic vectors $T^c$ and vision features $I_{ij}$ are calculated as $\mathbf{S}^{human}$. In Figure \ref{SIS}(c), given foreground mask $\mathbf{M}^{t}$, the prediction $\mathbf{M}$ is decided by the overlapping with $\mathbf{S}^{human}$.

\subsection{Geometric-aware Volume-based Detector}

Both the manual BEV annotations and pedestrian masks as 2D pseudo labels from SIS can be regarded as different observations of the 3D pedestrian density, i.e., the former are the top-down observations, and the latter the surroundings. 

To learn from these multi-view pseudo labels and predict the BEV pedestrian positions, a 3D volume as a ``bridge'' between them is constructed from encoding the input multi-view images by our proposed fully-unsupervised detector. 

\subsubsection{2D-to-3D Geometric Projection for 3D Volume}

Inspired by the previous multi-view methods \cite{rukhovich2022imvoxelnet, tu2023imgeonet}, we extract 2D features $\mathbf{F}^n \in \mathbb{R}^\mathrm{C\times H \times W}$ of each view $n$ via a visual encoder \cite{he2016deep} as is shown in Figure \ref{GVD}(a), where $\mathrm{C}$ is the channel number. Then, pixel-wise feature $F^n_{uv} \in \mathbb{R}^\mathrm{C}$ is back-projected into potential 3D voxels via 3D geometry. 

Formally, given a pinhole camera calibrated with intrinsic and extrinsic matrices $\{\mathbf{K}^n, \mathbf{\Psi}^n\}$, its geometric model to capture a pixel $(u, v)^\top$ from a 3D voxel $(x, y, z)^\top$  is: 
\begin{equation}
	\label{Backproj}
	\begin{bmatrix}
		u \\
		v \\
		1
	\end{bmatrix}
	= \frac{1}{\lambda} \mathbf{K}^n \mathbf{\Pi}^0 \mathbf{\Psi}^n
	\begin{bmatrix}
		x \\
		y \\
		z \\
		1
	\end{bmatrix}, 
	\mathbf{\Pi}^0 =
	\begin{bmatrix}
		1&0&0&0 \\
		0&1&0&0 \\
		0&0&1&0
	\end{bmatrix},
\end{equation}
where $\mathbf{\Pi}^0$ is an auxiliary matrix to obtain $(x', y', z')^\top$ from the 3D homogeneous coordinate $(x',y',z',1)^\top$ after the transformation by extrinsic matrix $\mathbf{\Psi}^n$, and $\lambda$ is the depth distance along optical axis between the center of this voxel and camera. $\frac{1}{\lambda}$ ensures the 2D homogeneity of $(u, v, 1)^\top$.

In 3D geometry, there are various 3D voxel $(x,y,z)^\top$ with different depths $\lambda$ via Eq. \ref{Backproj}, that can derive the same 2D pixel position $(u, v, 1)^\top$. For such a one-to-many 2D-3D correspondence, the 2D pixel feature $F^n_{uv}$ is assigned to all these 3D voxel positions to form a 3D volume $\mathbf{V}^n \in \mathbb{R}^\mathrm{C\times X\times Y\times Z}$ of the features from each camera view $n$: 
\begin{equation}
	\label{Assign}
	\mathbf{V}^n[:, x, y, z] =\mathbf{F}^n[:, u, v],
\end{equation}
where the 3D voxel coordinate $(x, y, z)^\top$ is from a subset of all voxels, which satisfies the one-to-many 2D-3D correspondence in Eq. \ref{Backproj} with the same 2D pixel position $(u, v)^\top$. 

In order to fuse the 3D volumes $\mathbf{V}^n$ in Figure \ref{GVD}(b) into a unified one $\mathbf{V}$, a soft-max function is employed as a relaxed version of maximum following other 3D volume-based methods \cite{iskakov2019learnable, nguyen2023deep} to re-weight each volume. To predict the 3D density $\boldsymbol{\mathcal{D}} \in [0, 1]^\mathrm{X\times Y\times Z}$ and color $\boldsymbol{\mathcal{C}} \in [0,1]^\mathrm{3\times X\times Y\times Z}$ from $\mathbf{V}$, a 3D-convolution network is adopted as a decoder, following the widely-used 2D-convolution of the previous methods \cite{hou2020multiview, song2021stacked, hou2021multiview, qiu20223d, engilberge2023two} on the BEV features.

\begin{table*}
	\caption{Detailed information of the multi-view pedestrian detection datasets for performance evaluation.}
	\label{Datasets}
	\centering
	\begin{tabular}{l|clccccc}
		\hline
		Datasets & 
		\begin{tabular}[c]{@{}l@{}}Camera \\ Number\end{tabular} &
		\begin{tabular}[c]{@{}l@{}}Input \\ Resolution\end{tabular} &
		\begin{tabular}[c]{@{}l@{}}Data \\ Collection\end{tabular} & \begin{tabular}[c]{@{}l@{}}Train\\ Frames\end{tabular} & 
		\begin{tabular}[c]{@{}l@{}}Test\\ Frames\end{tabular} & 
		\begin{tabular}[c]{@{}l@{}}Area\\ ($m\times m$)\end{tabular} &
		\begin{tabular}[c]{@{}l@{}}Crowdedness\\ (person/frame)\end{tabular}
		\\
		\hline
		Wildtrack \cite{chavdarova2018wildtrack} & 7& $1920\times1080$  & Real World & 360    & 40   & $12\times36$ &20\\
		Terrace \cite{fleuret2007multicamera} & 4 & $360\times288$  & Real World & 300    & 200   & $5.3\times5$&20\\
		MultiviewX \cite{hou2020multiview} & 6 & $1920\times1080$  & Simulation & 360    & 40   & $16\times25$&40\\
		\hline
	\end{tabular}
\end{table*}

\subsubsection{3D-to-2D Rendering Losses with Pseudo Labels}

With the 2D pedestrian masks $\mathbf{M}$ as pseudo labels, differential rendering framework PyTorch3D \cite{johnson2020accelerating} renders the predicted 3D density $\boldsymbol{\mathcal{D}}$ into 2D masks $\mathbf{\tilde{M}}$ by the camera model in Eq. \ref{Backproj}. Similarly, to discriminate crowded pedestrians by their appearances, the predicted 3D colors $\boldsymbol{\mathcal{C}}$ are rendered into 2D images $\mathbf{\tilde{I}}$. Since only the high-density parts have colors, the pseudo labels are input images $\mathbf{I}$ masked by $\mathbf{M}$. Following the instruction of PyTorch3D \cite{johnson2020accelerating}, Huber Loss \cite{huber1992robust} $\mathcal{L}_{huber}$ is used between these predictions and labels.

\subsection{Vertical-aware BEV Regularization}

As is shown in Figure \ref{Pipeline}(c) and \ref{GVD}(d), if predicted pedestrians are leaning or laying down, their BEV occupancy, i.e., maximized 3D density along Z-axis, is unnaturally larger than standing. Thus, Vertical-aware BEV Regularization (VBR) is proposed to regularize the vertical poses of pedestrians: 
\begin{equation}
	\mathcal{L}_\mathit{VBR}(\boldsymbol{\mathcal{D}}) = \frac{1}{\mathrm{XY}} \sum_{x, y=1}^{\mathrm{XY}} \big\vert \max(\mathcal{D}_{xyz})_z \big\vert, \mathcal{D}_{xyz} \in \boldsymbol{\mathcal{D}}.
\end{equation} 

Finally, the overall loss function $\mathcal{L}$ to jointly optimize our UMPD to predict the best 3D color and density $\{\boldsymbol{\mathcal{C}}^*, \boldsymbol{\mathcal{D}}^*\} = \arg\min\mathcal{L}$ via the rendered $\mathbf{\tilde{I}}$ and $\mathbf{\tilde{M}}$ is formulated as:
\begin{equation}
	\begin{split}
		\mathcal{L} & =  \mathcal{L}_\mathit{color} + \mathcal{L}_\mathit{mask} + \mathcal{L}_\mathit{VBR} 
		\\ & = \mathcal{L}_{huber}(\mathbf{\tilde{I}}, \mathbf{I}) + \mathcal{L}_{huber}(\mathbf{\tilde{M}}, \mathbf{M}) + \mathcal{L}_\mathit{VBR}(\boldsymbol{\mathcal{D}}).
	\end{split}
\end{equation}

For inference, the detection results of our proposed UMPD are derived by the predicted 3D density projected on BEV, i.e., $\mathbf{\Omega} = {\max(\mathcal{D}_{xyz})_z} \in \mathbb{R}^\mathrm{X\times Y}$, where $\mathcal{D}_{xyz} \in \boldsymbol{\mathcal{D}}$.

\section{Experiments}

In this section, extensive experiments are conducted on popular multi-view pedestrian detection benchmarks, i.e., Wildtrack, Terrace, and MultiviewX, to evaluate our proposed UMPD. Ablation study is performed on the key components SIS, GVD, and VBR. Qualitative analysis and state-of-the-art comparisons on these benchmarks are also reported.

\subsection{Datasets}

Wildtrack \cite{chavdarova2018wildtrack} and Terrace \cite{fleuret2007multicamera} datasets are collected from real-world interested regions surrounded by multiple calibrated and synchronized cameras, where the unscripted pedestrians are naturally walking or standing. MultiviewX \cite{hou2020multiview} is a challenging dataset using Unity 3D Engine to simulate more populated scenes, featured with $2\times$ higher crowdedness. Multiple Object Detection Accuracy and Precision \cite{kasturi2008framework} (MODA and MODP), Precision, and Recall are evaluation metrics. Table \ref{Datasets} shows more details of these datasets. 

\subsection{Implementation Details}

Our proposed UMPD method is based on PyTorch \cite{paszke2019pytorch} framework. Pretrained features in SIS are extracted by an unsupervised model \cite{oquab2023dinov2,caron2021emerging, caron2020unsupervised, grill2020bootstrap}. $\mathbf{S}^\mathit{human}$ is generated by ResNet \cite{he2016deep} of CLIP \cite{radford2021learning, zhou2022extract}. $4\times$A5000 GPUs are used for training, and $1\times$GPU for testing. Our SIS and VBR are only for training, and the inference time of GVD is $\mathtt{\sim}1.0$s/frame.

\begin{table}[t]
	\centering
	\caption{Ablation study on our proposed Semantic-aware Iterative Segmentation (SIS) about the PCA iteration T\textsubscript{PCA} and the usage of zero-shot semantic capability from CLIP to identify pedestrians.}
	\begin{tabular}{cc|cccc}
		\hline
		T\textsubscript{PCA}       & CLIP           & MODA          & MODP          & Precision   & Recall       \\
		\hline
		\multirow{2}{*}{1}
		& &15.0 &58.0 &95.0 &15.9 \\
		&\checkmark &19.1 &55.1 &68.9 &34.9 \\
		\cline{1-6}
		\multirow{2}{*}{2}
		& & 49.8 &58.5 &89.5 &56.4 \\
		&\checkmark &\textbf{76.6} &\textbf{61.2} & 90.1 & 86.0 \\			
		\cline{1-6}
		\multirow{2}{*}{3}
		& & 9.1 &43.7 &85.0 &11.0 \\
		&\checkmark &57.4 &60.9 &91.4 &63.3 \\
		\hline
	\end{tabular}
	\label{Ablation-SIS}
\end{table}

\subsection{Ablation Study}

The ablation study is performed on the popular real-world dataset Wildtrack. In Table \ref{Ablation-SIS}, for our proposed SIS, different PCA iteration T\textsubscript{PCA} and the zero-shot semantic capability of CLIP are experimented. ``T\textsubscript{PCA}=1'' performs worse with coarse-grained foreground as shown in Figure \ref{SIS}(a). Meanwhile, ``T\textsubscript{PCA}=3'' causes over-segmentation. Although CLIP fixes some mistakes by keeping some pedestrians parts as the foregrounds, the results are still not ideal. Equipped with CLIP to identify pedestrians and ``T\textsubscript{PCA}=2'', our UMPD performs the best. Finally, our proposed SIS adopts the powerful vision-language model CLIP to identify pedestrians as well as proper PCA iterations. 

Meanwhile, Table \ref{Ablation-GVD} evaluates our proposed GVD and VBR by: 1) 2D-to-3D geometric projection with different volume fusion; 2) 3D-to-2D rendering losses such as $\mathcal{L}_{\mathit{color}}$ and $\mathcal{L}_{\mathit{VBR}}$, since $\mathcal{L}_{\mathit{mask}}$ is inevitable to predict detection results $\mathbf{\Omega}$. For the volume fusion operations, the soft-max re-weighting achieves higher MODA and MODP, while simpler adding and concatenation are insufficient to directly handle such complex 3D volumes. 

For the loss functions, $\mathcal{L}_{\mathit{color}}$ assists the detection by discriminate the different appearances of pedestrian instances with higher MODA and MODP than merely $\mathcal{L}_{\mathit{mask}}$ used. Note that our $\mathcal{L}_{\mathit{VBR}}$ follows the natural vertical poses of humans and thus brings significant improvements (+49.3 MODA and +9.1 MODP) on the default loss functions $\mathcal{L}_{\mathit{mask}}$ and $\mathcal{L}_{\mathit{color}}$ from the instruction of PyTorch3D \cite{johnson2020accelerating}. 

\begin{table}[t]
	\centering
	\caption{Ablation study on our proposed Geometric-aware Volume-based Detector (GVD) and Vertical-aware BEV Regularization (VBR). Concatenation and adding fusions are compared with soft-max. Loss functions $\mathcal{L}_{\mathit{color}}$ and $\mathcal{L}_{\mathit{VBR}}$ are also ablated.}
	\begin{tabular}{l|cccc}
		\hline
		Settings           & MODA          & MODP          & Precision   & Recall       \\
		\hline
		\textbf{our UMPD} &\textbf{76.6} &\textbf{61.2} & 90.1 & 86.0 \\
		\hline
		Soft. $\rightarrow$ Con. &71.3 &60.8 &83.4 &89.1 \\
		Soft. $\rightarrow$ Add. &74.5 &61.0 &86.1 &88.9 \\
		\hline
		$\rightarrow$ w/o $\mathcal{L}_{\mathit{color}}$ &71.0 &60.3 &83.2 &89.0 \\
		$\rightarrow$ w/o $\mathcal{L}_{\mathit{VBR}}$ &27.3 &52.1 &65.9 &56.7 \\
		\hline
	\end{tabular}
	\label{Ablation-GVD}
\end{table}

\begin{table}[t]
	\centering
	\caption{Different segmentation methods for 2D masks as pseudo labels to train our UMPD: our proposed method SIS, unsupervised method CutLER \cite{wang2023cut}, and supervised method Mask2Former \cite{cheng2022masked}. }
	\begin{tabular}{l|cccc}
		\hline
		Methods           & MODA          & MODP          & Precision   & Recall       \\
		\hline
		\begin{tabular}[c]{@{}l@{}}SIS(=\textbf{UMPD}) \\ \textit{unsupervised}\\ \end{tabular} &76.6 &61.2 & 90.1 & 86.0 \\
		\hline
		\begin{tabular}[c]{@{}l@{}}$\rightarrow$ CutLER \\ \textit{unsupervised}\\ \end{tabular} &38.9 &53.5 &80.5&51.3 \\
		\begin{tabular}[c]{@{}l@{}}$\rightarrow$ Mask2For. \\ \textit{supervised}\\ \end{tabular} &79.3 &63.3 &90.1 &89.1 \\
		\hline
	\end{tabular}
	\label{Segment}
\end{table}

\subsection{Different Segmentation Methods for 2D Masks}

\begin{figure*}
	\centering
	\includegraphics[width=1.0\textwidth]{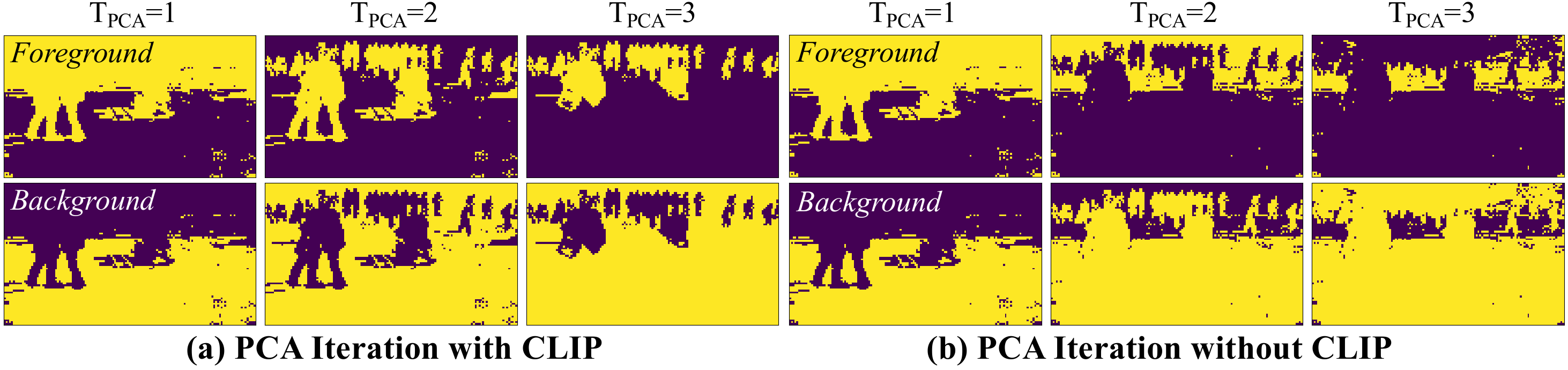}
	\caption{Visualization the 2D masks generated by our proposed SIS with different T\textsubscript{PCA} and the zero-shot semantic capability of CLIP. Naïve increasing the T\textsubscript{PCA} leads to wrong decision via $> \phi_\mathit{thr}$ as foreground in Eq. \ref{PCA-naive}, while CLIP identifies the pedestrian parts correctly. }
	\label{PCA-Vis}
\end{figure*}

\begin{figure*}
	\centering
	\includegraphics[width=1.0\textwidth]{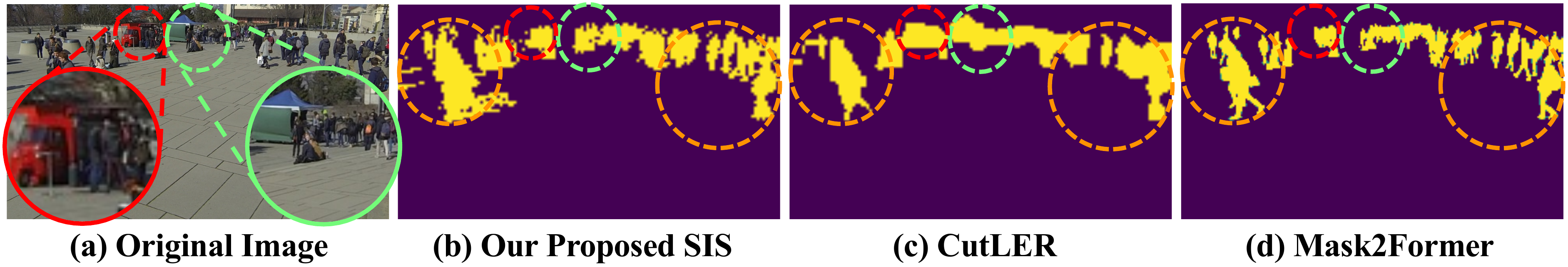}
	\caption{Results of different segmentation methods. CutLER \cite{wang2023cut} segments all salient objects like the circled red truck and green tent. Supervised Mask2Former \cite{cheng2022masked} method yields more fine-grained masks in orange circles than those of the unsupervised methods including CutLER and our proposed SIS, which serves as a pseudo label generator for our proposed UMPD to gain better performance  in Table \ref{Segment}.}
	\label{Segment-Vis}
\end{figure*}

As a mainstream unsupervised segmentation methods, CutLER \cite{wang2023cut} is different from our proposed SIS, which can only segment all objects that are salient in a single image, rather than those of inter-image concepts, e.g., pedestrians in multi-view images. Meanwhile, another powerful method Mask2Former \cite{cheng2022masked} is initialized with unsupervised model \cite{oquab2023dinov2,caron2021emerging, caron2020unsupervised, grill2020bootstrap}, and fine-tuned via supervised learning. 

In Table \ref{Segment}, we compare the 2D segmentation masks from these different unsupervised or supervised methods, as pseudo labels to train our UMPD. Noisy masks with non-human objects from CutLER leads to lower performance. Mask2Former yields fine-grained masks and better results by its supervised fine-tuning, while our proposed fully-unsupervised SIS also performs competitively.

\subsection{Qualitative Analysis}

In Figure \ref{PCA-Vis}, we firstly visualize the unsupervised 2D masks generated by our proposed SIS with different PCA iteration T\textsubscript{PCA} and whether to use the zero-shot semantic capability of CLIP. At T\textsubscript{PCA}=1, foregrounds in Figure \ref{PCA-Vis}(a) and (b) are coarse-grained with pedestrians and other objects, which is handled by the CLIP and $>\phi_{thr}$ part in Eq. \ref{PCA-naive} (w/o CLIP), respectively. At T\textsubscript{PCA}=2, the foreground is correctly identified by CLIP in Figure \ref{PCA-Vis}(a), while the $>\phi_{thr}$ part yields wrong mask, since only inter-image concepts are segmented, regardless of the foreground semantic class. Under the over-segmentation at T\textsubscript{PCA}=3, CLIP decreases some mistakes, which are consistent with the results in Table \ref{Ablation-SIS}.

In the meantime, 2D masks by different segmentation methods compared in Table \ref{Segment} are shown in Figure \ref{Segment-Vis}. Noisy masks by CutLER \cite{wang2023cut} comprises some non-human salient objects like tent and truck, which misleads the detector during training and leads to worse performances in Table \ref{Segment}. With the more powerful supervised fine-tuning, Mask2Former \cite{cheng2022masked} yields more fine-grained masks than our unsupervised SIS, e.g., the pedestrians with various scales and crowdednesses in the orange circles of Figure \ref{Segment-Vis}(d). 

\begin{figure*}
	\centering
	\includegraphics[width=1.0\textwidth]{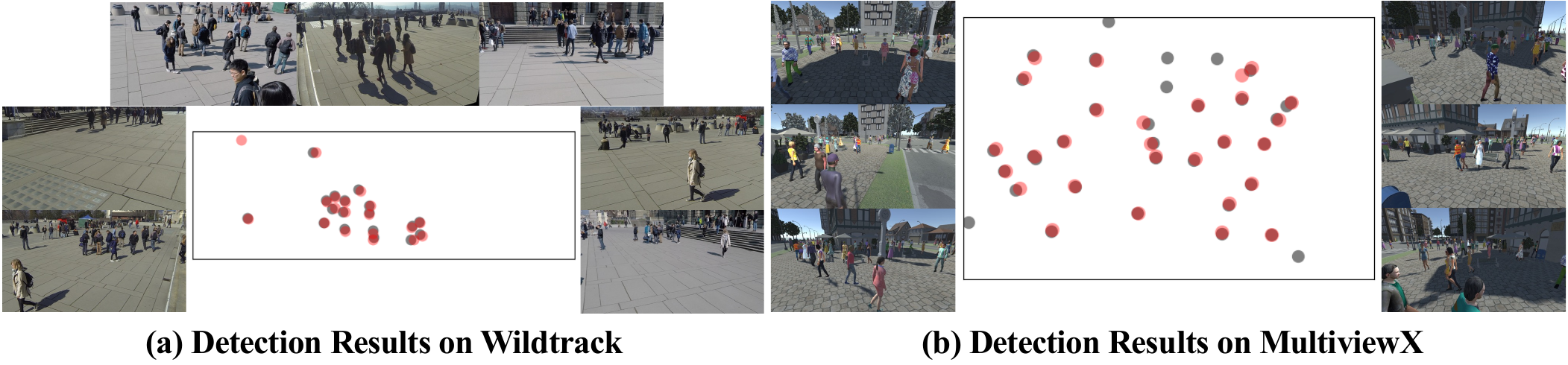}
	\caption{Detection results by our proposed UMPD on Wildtrack \cite{chavdarova2018wildtrack} and MultiviewX \cite{hou2020multiview} datasets. Gray and Red dots are ground truths and the predictions from UMPD, respectively. (a) Even though the crowdedness of Wildtrack in Table \ref{Datasets} is lower, some pedestrians are more heavily occluded and still detected by our UMPD. (b) MultiviewX brings more challenges via larger-scale and more populated scenes. }
	\label{Detect-Vis}
\end{figure*}

\begin{table*}
	\centering
	\caption{Comparisons with the state-of-the-arts on multi-view pedestrian detection datasets. Note that our proposed UMPD is unsupervised.}
	\begin{tabular}{c|cccc|cccc|cccc}
		\hline
		\rule{0pt}{8pt}
		& \multicolumn{4}{c|}{Wildtrack}               & \multicolumn{4}{c|}{Terrace}               & \multicolumn{4}{c}{MultiviewX}                               \\ 
		\cline{2-13}
		Methods                  & MODA          & MODP          & Pre.   & Rec.        & MODA          & MODP          & Pre.     & Rec.        & MODA          & MODP          & Pre.     & Rec. \\
		\hline
		RCNN \& Clu.     & 11.3          & 18.4          & 68          & 43      & -11		&28		&39		&50	      & 18.7          & 46.4          & 63.5         & 43.9         \\
		POM-CNN  & 23.2          & 30.5          & 75          & 55       &58		&46			&80		&78	      & -             & -             & -             & -             \\
		DeepMCD   & 67.8          & 64.2          & 85          & 82       &-			&-			&-			&-       & 70.0          & 73.0          & 85.7         & 83.3          \\
		Deep-Occlus. & 74.1        & 53.8          & 95 & 80      &71			&48		&88		&82      & 75.2         & 54.7          & 97.8 & 80.2          \\
		MVDet& 88.2 			& 75.7 			& 94.7        & 93.6 	&87.2		&70.0		&98.2		&88.8	& 83.9 		& 79.6 		& 96.8          & 86.7     \\
		SHOT & 90.2 		& 76.5 			& 96.1      & 94.0		&87.1		&70.3		&98.9		&88.1		& 88.3 		& 82.0 		& 96.6      & 91.5 \\
		MVDeTr  &91.5			&82.1			&97.4		&94.0		&-			&-			&-			&-	& 93.7		&91.3		&99.5	&94.2	\\
		3DROM	& 93.5			& 75.9 			& 97.2      & 96.2		&94.8		&70.5		&99.7		&95.1	 & 95.0 	& 84.9 		& 99.0      &96.1 \\
		MVAug &93.2		&79.8		&96.3		&97.0	&-			&-			&-			&-	&95.3		&89.7		&99.4		&95.9	\\
		\hline
		\textbf{UMPD (ours)} &76.6 &61.2 & 90.1 & 86.0 		&73.3	&62.5	&83.6	&91.3			&67.5   	&79.4		&93.4	&72.6			\\
		\hline
	\end{tabular}
	\label{SOTA}
\end{table*}

Moreover, Figure \ref{Detect-Vis} shows the detection results by our proposed UMPD method on Wildtrack \cite{chavdarova2018wildtrack} and MultiviewX \cite{hou2020multiview} datasets. Without any manual labels, UMPD detects the pedestrians correctly in both real-world and simulated scenes. There are still some wrong results near the edges of region, where the information from less overlapped camera views is insufficient for accurate detection, particularly in more populated MultiviewX like Figure \ref{Detect-Vis}(b).

\subsection{Comparisons with the State-of-the-arts}

In Table \ref{SOTA}, we compare our proposed unsupervised approach UMPD with fully-supervised state-of-the-art methods on Wildtrack, Terrace, and MultiviewX benchmarks: detection-based methods RCNN \& Clustering \cite{xu2016multi} and POM-CNN \cite{fleuret2007multicamera}; anchor-based methods DeepMCD \cite{chavdarova2017deep} and Deep-Occlusion \cite{baque2017deep}; perspective-based anchor-free methods MVDet \cite{hou2020multiview}, SHOT \cite{song2021stacked}, MVDeTr \cite{hou2021multiview}, 3DROM \cite{qiu20223d}, and MVAug \cite{engilberge2023two}. For Wildtrack and Terrace, our proposed UMPD surpasses RCNN \& Clustering, POM-CNN, Deep-Occlusion by both MODA and MODP, and DeepMCD by MODA. As is shown in Table \ref{Datasets} and Figure \ref{Detect-Vis}(b), MultiviewX is featured with $2\times$ higher crowdedness. On this challenging benchmark, our UMPD out-performs RCNN \& Clustering by MODA and MODP, as well as DeepMCD and Deep-Occlusion by MODP. In conclusion, our fully-unsupervised UMPD achieves competitive performances on all real-world and simulated datasets, without any manual annotations essential to mainstream supervised methods.

\section{Conclusion}

In this paper, we have proposed a novel unsupervised multi-view pedestrian detection approach UMPD, which eliminates the heavy burden of manual annotations. For such a challenging task, three key components are proposed: SIS segments the PCA values of pretrained features iteratively into 2D masks, with zero-shot semantic capability of CLIP to identify pedestrians. GVD encodes multi-view images into a 3D volume, and learns to predict density and color by the rendering losses with SIS masks as labels. VBR constraints the 3D density predicted by GVD to be vertical on the BEV ground plane, following the natural poses of most pedestrians. With these components, our proposed UMPD achieves competitive performances on challenging benchmarks Wildtrack, Terrace, and MultiviewX. We hope this work, as the first fully-unsupervised method in this field, could be a start and inspire more interesting future works.

{
    \small
    \bibliographystyle{ieeenat_fullname}
    \bibliography{main}
}


\end{document}